\documentclass[letterpaper, 10 pt, conference]{ieeeconf}  

\IEEEoverridecommandlockouts                              

\usepackage[utf8]{inputenc} 
\usepackage[T1]{fontenc}    
\usepackage{hyperref}       
\usepackage{url}            
\usepackage{booktabs}       
\usepackage{amsfonts}       
\usepackage{nicefrac}       
\usepackage{microtype}      
\usepackage{lipsum}
\usepackage{algorithm,algcompatible}
\usepackage{comment} 
\usepackage{lipsum} 
\usepackage{graphics} 
\usepackage{epsfig} 
\usepackage{times} 
\usepackage{amsmath} 
\usepackage{amssymb}  
\usepackage{color}
\usepackage{xcolor}
\usepackage{caption}
\usepackage{mathabx}
\newcommand{\norm}[1]{\left\lVert#1\right\rVert}
\newtheorem{theorem}{Theorem}[section]

\newtheorem{definition}[theorem]{Definition}

\newtheorem{assumption}[theorem]{Assumption}
\usepackage{algorithm}
\usepackage{algorithmicx}
\usepackage{algpseudocode}
\usepackage{makecell}

\title{\LARGE \bf
Model Predictive Instantaneous Safety Metric for \\Evaluation of Automated Driving Systems
}

\author{Bowen Weng$^{1}$, Sughosh J. Rao$^{1}$, Eeshan Deosthale$^{1}$, Scott Schnelle$^{2}$ and Frank Barickman$^{2}$
\thanks{$^{1}$Bowen Weng ({\tt\small bowen.weng.ctr@dot.gov}), Sughosh J. Rao ({\tt\small sughosh.rao.ctr@dot.gov}) and Eeshan Deosthale ({\tt\small deosthale.2@osu.edu}) are with the Transportation Research Center Inc., East Liberty, Ohio 43319, USA}%
\thanks{$^{2}$Scott Schnelle ({\tt\small scott.schnelle@dot.gov}) and Frank Barickman ({\tt\small frank.barickman@dot.gov}) are with the National Highway Traffic Safety Administration, USA}%
}

\begin{document}

\maketitle
\thispagestyle{empty}
\pagestyle{empty}

\begin{abstract}
Vehicles with Automated Driving Systems (ADS) operate in a high-dimensional continuous system with multi-agent interactions. This continuous system features various types of traffic agents (non-homogeneous) governed by continuous-motion ordinary differential equations (differential-drive). Each agent makes decisions independently that may lead to conflicts with the subject vehicle (SV), as well as other participants (non-cooperative). 
A typical vehicle safety evaluation procedure that uses various safety-critical scenarios and observes resultant collisions (or near collisions), is not sufficient enough to evaluate the performance of the ADS in terms of operational safety status maintenance. In this paper, we introduce a \emph{Model Predictive Instantaneous Safety Metric} (MPrISM), which determines the safety status of the SV, considering the worst-case safety scenario for a given traffic snapshot. The method then analyzes the SV's closeness to a potential collision within a certain evaluation time period. The described metric induces theoretical guarantees of safety in terms of the time to collision under standard assumptions. Through formulating the solution as a series of minimax quadratic optimization problems of a specific structure, the method is tractable for real-time safety evaluation applications. Its capabilities are demonstrated with synthesized examples and cases derived from real-world tests.
\end{abstract}


\section{INTRODUCTION}
Automated Driving Systems (ADSs) operate in a high-dimensional state space with various types of other traffic agents. While the instantaneous states of each agent, including location, velocity and the type of agent may be accessible through perfect perception and communication, the actual motion trajectory carried out by each agent is largely unknown and occasionally random. A pedestrian could cross the street without the presence of a green traffic light or cross walk~\cite{Wikipedia.org2018Death}, a lead-vehicle could perform abrupt maximum braking for collision avoidance, people could start chasing a bird in the middle of the road~\cite{Guardian2017}, to name a few. 

For the intelligent subject vehicle (SV) to safely navigate in such a complex environment, it must maintain a certain level of control over the outcome even when presented with the worst possible scenario. A way to evaluate safety performance, then, is through exposing the SV to various safety-critical scenarios and observing the resultant consequences, e.g., whether a collision occurs. Conflict severity based on the instantaneous vehicle states like Time to Collision (TTC)~\cite{lee1976theory} is also included when applicable. Historically, TTC is the dominant metric in analyzing the safety of intelligent longitudinal control systems~\cite{barickman2009nhtsa, forkenbrock2015nhtsa}. Notice that some variants of TTC also incorporate probabilistic risk assessment~\cite{schreier2017bayesian}, and consider the capability of SV in avoiding potential collisions by incorporating extreme evasive maneuvers~\cite{minderhoud2001extended}. 

However, the safety evaluation of ADS fundamentally presents two unique challenges to the described process. 
First, for safety assessment within a high-dimensional continuous domain, simply observing collisions is no longer sufficient. Even if no collision is encountered, a highly automated driving algorithm can still decide to drive aggressively through crowded traffic, experiencing various near-miss scenarios. An adequate metric for safety evaluation and conflict assessment should be able to distinguish such a risky algorithm from conservative ones analytically. 
Second, the ADS is capable of operating both longitudinally and laterally with intelligent decision making and control. Previous analyses, such as TTC, are restricted to a small subset of all possible scenarios. 
On the other hand, if we consider the full actuation capabilities within the complete Euclidean operable domain, it would lead to a complex strategy for both the traffic agent to initiate a dangerous situation and the SV to guarantee its escape correspondingly. Only considering extreme longitudinal evasive maneuvers (e.g., maximum braking) is no longer sufficient to justify the safety status of the SV over the potential roadway area occupied in multi-agent interactions.

In this paper, we formulate the safety evaluation problem within a general multi-agent system of various types of agents (non-homogeneous) following independent control and planning strategies (non-cooperative) with continuous motion governed by ordinary differential equations (differential-drive). For an arbitrary agent selected as the SV, we introduce the \emph{Model Predictive Instantaneous Safety Metric} (MPrISM) that seeks to determine its instantaneous capability of getting away from a potential collision under standard assumptions. It can also be interpreted as a high-dimensional model predictive TTC metric. The metric considers the full control authority of the SV and the traffic agent, along with game interaction between them to estimate the time-to-collision when the traffic agent takes the most aggressive action to cause a crash as the SV tries its best to avoid it. This is in contrast to the classic TTC calculation considering only the longitudinal states of all traffic participants with the strong assumption of all agents maintaining the current velocity and orientation for all time. 
The described method is particularly effective in the safety evaluation of ADS given its analytical form of solving a series of minimax quadratic optimization problems subject to linear constraints. The metric is implemented in simulated and real-world traffic scenarios incorporating vehicles of car-like kinematics, as well as pedestrians of single-integration kinematics.

\subsection{Related Work}
The definition of multi-agent safety is dependent upon the cooperative pattern of all participating agents~\cite{bowen2020presentation}. If one assumes the agents are all absolutely cooperative in avoiding potential conflicts, the optimal reciprocal collision avoidance (ORCA)~\cite{van2011reciprocal} method and its extensions\cite{van2011reciprocalacc, bareiss2015generalized} provide the safety guarantees for an arbitrary differential-drive agent operating in a non-homogeneous multi-agent system, with all agents following a consensus strategy. For on-road vehicle operation, traffic rules and responsibility recognition can also be considered as a consensus strategy to achieve cooperative collision avoidance. On the other hand, if all agents are independent in making planning and control decisions, the safety evaluation problem under such a non-cooperative assumption is closely related to reachability analysis (RA) originated from differential game and optimal control theory. RA provides theoretical guarantees for the safety properties of nonlinear control systems~\cite{mitchell2007comparing}. It has been widely adopted in designing safety-critical motion planning algorithms in various differential games~\cite{grote1975theory, sun2017pursuit}, as well as mobile robots~\cite{gillula2010design} in engineering practice. In particular, Althoff et al.~\cite{althoff2014online} have studied the safe operable space of ADS through propagating the forward reachable set with various constraints. The concept has been further extended to various sub-problems within the research domain of ADS safety~\cite{klischat2019generating}. However, in general, applying RA in analyzing the safety properties of ADS is computationally intensive and would require strict assumptions as well as various approximations to work in even simple scenarios.

For the particular problem of automated vehicle safety evaluation in practice, various methods have been proposed with real-time computational efficiency. The responsibility-sensitive safety (RSS) metric~\cite{shalev2017formal} is conceptually close to the approximation of the backward reachability set (BRS). A series of scenario-specific rules are established with respect to various perspectives in real-life driving, including responsibility, traffic laws, right-of-way rules, and object visibility. This essentially enables the method in practice to approximate a safety bound of actions within which the SV is safe. However, the method significantly simplifies the motion kinematics and is not a continuous metric regarding the safety status of the SV. Another similar approach is the safety force field~\cite{nvidia2019}, which conceptually approximates the control barrier. Junietz et al.~\cite{junietz2018criticality} recently proposed a model-based safety evaluation method for automated vehicles by selecting a target function representing the criticality (safety) level of a potential trajectory. The safety evaluation task then becomes a trajectory optimization problem, specifically, a quadratic programming problem with linear constraints, which comes with tractable solutions. However, the safety analysis of such form can become very sensitive to how the safety criterion is determined. The notion of criticality is only monotonic with respect to a certain safety status induced by the design of the target function. It does not necessarily indicate the rigorous closeness to a potential collision. 

\subsection{Organization}
Section~\ref{sec:problem} provides preliminaries for our method, including the analytic definition of safety used for this metric. Section~\ref{sec: mpis} formally introduces MPrISM and details of MPrISM for ADS safety evaluation. Various examples showing the capability and effectiveness of MPrISM are presented in Section~\ref{sec:examples}. Section~\ref{sec:conclusion} summarizes the paper and provides some considerations for possible future work.

\section{PROBLEM FORMULATION}
\label{sec:problem}
We formulate the general problem of ADS safety analysis in a non-homogeneous differential-drive multi-agent system with partially non-cooperative traffic agents. 
 
\subsection{The Differential-drive Agents}
Consider the general motion of an intelligent agent denoted by the subscript $i$ satisfying
\begin{equation}
    \label{eq:motion ode}
    \dot{\mathbf{x}_i} = f_i(\mathbf{x}_i,\mathbf{u}_i,t),
\end{equation}
where the agent state $\mathbf{x}_i \in \mathcal{X}_i \subseteq \mathbb{R}^{N(i)}$, action $\mathbf{u}_i \in \mathcal{U}_i \subseteq \mathbb{R}^{M(i)}$ and the function $f_i : \mathcal{X}_i \times \mathcal{U}_i \times \mathbb{R} \rightarrow \mathcal{X}_i$. Given the multi-agent system is non-homogeneous, all agents do not necessarily share the same admissible actions space. Here we assume they all have the same dimensions of actions with $M(i) = M, \forall i$. We also assume all agents operate within the Euclidean plane, i.e., the state $\mathbf{x}_i$ contains at least the two-dimensional coordinates $(p_i,q_i)$ and $N(i) \geq 2$. In practice, both states and actions can be bounded and sometimes are of periodic dimension (e.g., heading angle), we consider $\mathbb{R}$ for simplification.

Presented with an initial condition $\mathbf{x}_i(0)$, let $\mathbf{x}_i = g_i(\mathbf{x}_i,\mathbf{u}_i, t)$ be the solution of~\eqref{eq:motion ode} where $g_i : \mathcal{X}_i \times \mathcal{U}_i \times \mathbb{R} \rightarrow \mathcal{X}_i$. Given an arbitrarily small step size $\Delta$, one can numerically approximate the solution through the Runge-Kutta method, i.e., 
\begin{equation}
    \label{eq:rk4 motion}
   \mathbf{x}'_i = g_i^{rk}(\mathbf{x}_i,\mathbf{u}_i,\Delta)
\end{equation}
where $x'_i$ approximates the state at $t=\Delta$ from executing action $u_i$ for state $x_i$ at $t=0$. A more detailed formulation of vehicle and pedestrian motion following~\eqref{eq:motion ode} and~\eqref{eq:rk4 motion} will be addressed in Section~\ref{sec: mpis}.

\subsection{The Definition of Safety}
Generally, relative distance determines collision and collision avoidance implies safety. Consider the system of multi-agent interactions among $k+1$ agents of various types having the state $\mathbf{x} = \{\mathbf{x}_i\}_{i=0,1,...,k}$ with $\mathbf{x}_i = [p_i, q_i]$ being the Euclidean position of the $i$-th agent. The index $0$ denotes the SV. The collision set is then defined as
\begin{equation}
    \label{eq:crash set}
    \Omega_{C} = \{\mathbf{x} \mid \underset{i = 1,2,...,k}{\inf} \{ d(\mathbf{x}_i, \mathbf{x}_0) \} \leq C \}.
\end{equation}
That is, a collision is determined to occur if and only if there exists at least one principal other agent whose relative distance measured through the metric $d(\cdot, \cdot)$ with respect to the test subject is smaller or equal to a given constant $C$. Throughout this paper, we adopt the Euclidean distance as the distance metric, i.e., $d(\mathbf{x}_i, \mathbf{x}_0) = \norm{\mathbf{x}_i-\mathbf{x}_0}_2$.

We also consider the following assumption that is commonly accepted in robotics motion planning~\cite{bareiss2015generalized} and is also valid in the engineering practice of automotive research.
\begin{assumption}
    \label{asp:fix delta}
    All participating agents share a common and constant sensing-action cycle that is the same as the selected step size $\Delta$ for numerical approximation of~\eqref{eq:motion ode}. Starting at time $t$, each agent selects a control action $\mathbf{u}(t) \in \mathcal{U}(t)$ from its admissible action space and consistently executes the same action throughout the time interval of $[t, t+\Delta)$. Furthermore, the admissible action space within such time interval also remains the same as $\mathcal{U}(t)$.
\end{assumption}

Given the approximated solution~\eqref{eq:rk4 motion}, we seek to justify the safety status of the SV by evaluating whether it maintains the capability of staying outside the collision set with appropriately selected trajectory of $T$ steps of actions $\bar{\mathbf{u}} = \bar{\mathbf{u}}(T) := [\mathbf{u}(0),\ldots,\mathbf{u}(T-1) ]$, $\bar{\mathbf{u}} \in \Pi_{n=1}^{T}\mathcal{U}_i(t+(n-1)\Delta)$ within time period of $T \Delta$ regardless of how other participating agents behave. 

\subsection{Partially Non-cooperative Collision Avoidance}
Generally, one has to assume all participating agents are non-cooperative, in order to provide absolute safety guarantees for navigating in the multi-agent system. However, as pointed out by various previous work~\cite{shalev2017formal}, such an assumption is often too strong to hold in practice. As illustrated in Fig.~\ref{fig:boxedin snap}, the SV becomes safety-critical easily in the boxed-in scenario if all surrounding vehicles decide to be aggressive. In this paper, we consider a more relaxed but also practical assumption of the \emph{partially non-cooperative} pattern as follows. Similar assumptions have been made in previous work without being specifically defined~\cite{shalev2017formal, chen2017obstacle}.
\begin{assumption}
    \label{asp:partial non-coop}
    Among $k$ principal other agents, there exists at most one non-cooperative agent and all other agents will comply with the SV to avoid collisions.
\end{assumption}
In the illustrated example from Fig.~\ref{fig:boxedin snap}, consider the case where the lead principal other vehicle (POV) performs an abrupt brake, enforcing the following SV to execute braking to escape. Assumption~\ref{asp:partial non-coop} states that the rear-POV will also brake to comply, and the two side-POVs will not initiate any trajectories that would make the situation worse. Notice that the SV does not hold any information regarding the identity of the non-cooperative agent other than the existence of such an agent among $k$ traffic objects. 

Given Assumption~\ref{asp:partial non-coop}, the SV can perform the safety check with respect to each traffic object individually and take the worst case among all results obtained. Let $h_i(\mathbf{x},\mathbf{u},t) = \norm{\mathbf{x}_i-\mathbf{x}_0}_2$ be the relative distance between the $i$-th traffic agent and the SV, $h(\mathbf{x}, \mathbf{u}, t) = \min_{i = 1,2,...,k} \{ h_i(\mathbf{x},\mathbf{u},t) \}$.
We conclude this section with the definition of safety in navigating in the non-homogeneous, differential drive, and partially non-cooperative multi-agent system. It will also shed light on the safety evaluation metric introduced in this paper.
\begin{definition}
    \label{def: safe plan alg}
    Consider the collision defined as~\eqref{eq:crash set} with assumptions of~\ref{asp:fix delta} and~\ref{asp:partial non-coop}. Let
    \begin{equation}
        \label{eq: minmax opt}
        h_i^*(\mathbf{x},\mathbf{u},t,T,\Delta) = \underset{\bar{\mathbf{u}}_i}{\min} \underset{\bar{\mathbf{u}}_0}{\max} \left( h_i(\mathbf{x},\mathbf{u},t+T\Delta) \right)
    \end{equation}
    The SV is defined as being safe with $\mathbf{x}(t) \notin \Omega_{C}$ within the time interval of $[t, t+T\Delta), T>0$ if $\forall i \in \{1,2,...,k\}, \forall n \in \{1,2,...,T\},$
    \begin{equation}
        \label{eq: safe plan}
        h_i^*(\mathbf{x},\mathbf{u},t,n,\Delta)  \geq C.
    \end{equation}
\end{definition}
Intuitively, the minimax optimization of~\eqref{eq: minmax opt} induces the optimal outcome available for the SV when presented with the worst scenario. If such an optimal outcome is not a collision (identified by~\eqref{eq: safe plan}), the SV is considered safe by the definition above. 
Generally, the optimization problem is computationally complex, considering the highly nonlinear motion of~\eqref{eq:motion ode} and the admissible action space $\mathcal{U}_i$ is often state-dependent and time-variant in engineering practice. However, as we will show in Section~\ref{sec: mpis}, through local linearization of various motion kinematics, one can achieve a tractable solution of~\eqref{eq: minmax opt} under standard assumptions, specifically for purposes of evaluating ADS safety performance broadly. 

\section{{Model Predictive Instantaneous Safety Metric}}
\label{sec: mpis}
We first describe the formal MPrISM algorithm. For the specific implementation of an ADS safety evaluation, through linearizing the vehicle and pedestrian motion kinematics, one can then simplify the proposed MPrISM algorithm to solving a series of minimax quadratic optimization problems under linear constraints.

\subsection{The Derived Method}
Given the system state $\mathbf{x}(t)$ at time $t$ consisting of one SV and $k$ traffic agents, referred to as a snapshot, Algorithm~\ref{alg:ttc} determines the notion of "time-to-collision" $\tau$ in the Euclidean operable domain.

\begin{algorithm}[b]
\vspace{0.3cm}
\caption{Model Predictive Time to Collision for a Given Snapshot} \label{alg:ttc} 
\begin{algorithmic}[1]
\State {\bf Input:} Snapshot $\mathbf{x}(t)$, step size $\Delta$, collision set defined as~\eqref{eq:crash set}, look-ahead steps $T$.
\For{ $i=1, \ldots, k$ }
 \State {Indicator of collision $c=False$}
 \For{ $j=1, \ldots, T$}
     \State {$\tau_i = j \Delta$}
     \If {$h_i^*(\mathbf{x},\mathbf{u},t,j,\Delta) \leq C$}
         \State $c=True$
         \State break
     \EndIf
 \EndFor 
 \If {$c$ is $False$}
\State $\tau_i = \tau_i + \Delta$
\EndIf
\EndFor
\State {{\bf Output:} $\tau = \underset{i=1, \ldots, k}{\min} \{ \tau_i \}$}
\end{algorithmic}
\end{algorithm} 

The derived $\tau$ from the above algorithm is referred to as the \emph{Model Predictive Time-to-Collision} (MPrTTC) in this paper. This is in contrast with the classic TTC~\cite{lee1976theory} where only the one-dimensional longitudinal motion is considered.

It is immediate that $\tau=T\Delta$ if the SV is collision-free within the time period of $[t, t+T \Delta)$. For $\tau < T \Delta$, $\tau$ represents the approximated time to collision of the SV when presented with the worst set of actions by the traffic agents going ahead from the current scenario snapshot. For the rest of this paper, we say the SV is safe if $\tau=T\Delta$ (i.e., no collision would occur per Definition~\ref{def: safe plan alg} within $T\Delta $ seconds in the future) and the SV is unsafe otherwise. 

Given a time series of $L$ snapshots, referred to as a scenario, one can perform various statistical propagations to analyze the overall performance of the SV in terms of safety. 

\subsection{Traffic Motion Kinematics}
We further address details of vehicle and pedestrian kinematics to help simplify the optimization of~\eqref{eq: minmax opt}. 

\subsubsection{Vehicle kinematics}
\begin{figure}
    \centering
    \vspace{0.2cm}
    \includegraphics[width=0.45\textwidth]{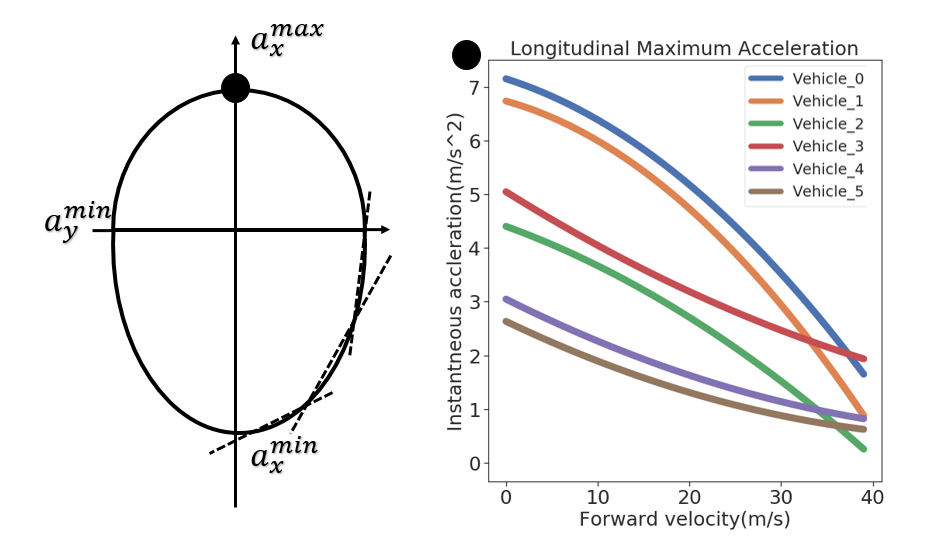}
    \caption{Approximating the admissible action space of vehicles with respect to the vehicle type and current speed. The black dot can be replaced by any of the values shown in the right subplot, given the vehicle type and the current speed.}
    \label{fig:a space veh}
\end{figure}

Given the nonlinear vehicle kinematics of
\begin{equation}
    \label{eq: vehicle kinematics}
    \begin{aligned}
    \dot{\mathbf{x}}\!=\!\begin{bmatrix} \dot{p} & \dot{q} & \dot{v} & \dot{\phi} \end{bmatrix}^T\!=\!\begin{bmatrix} v\cos{(\phi)} & v\sin{(\phi)} & a_x & \frac{a_y}{v} \end{bmatrix}^T,
    \end{aligned}
\end{equation}
with world coordinates $(p, q)$, velocity $v$, heading angle $\phi$ and control actions of $\mathbf{u} = [a_x, a_y]$. Assume small course angle and change of velocity, one can linearize the kinematics with a constant velocity $\tilde{v}$ as
\begin{equation}
    \label{eq: linear vehicle kinematics}
    \dot{\mathbf{x}} = \begin{bmatrix} \dot{p} & \dot{q} & \dot{v} & \dot{\phi} \end{bmatrix}^T \approx \begin{bmatrix} v & \tilde{v} \phi & a_x & \frac{a_y}{\tilde{v}} \end{bmatrix}^T.
\end{equation}
We also have the Runge-Kutta approximation with step size $\Delta$ of
\begin{equation}
    \label{eq: linear vehicle rk4}
    \begin{aligned}
        & \mathbf{x}' = \mathbf{A_v x} + \mathbf{B_v u}, \mathbf{x} = [p, q, v, \phi]^T, \mathbf{u}=[a_x, a_y]^T, \\
        & \mathbf{A_v} = \begin{bmatrix} 1 & 0 & \Delta & 0 \\
                                0 & 1 & 0 & \tilde{v} \Delta \\
                                0 & 0 & 1 & 0 \\
                                0 & 0 & 0 & 1\end{bmatrix},
          \mathbf{B_v} = \begin{bmatrix} \frac{\Delta^2}{2} & 0 \\
                                0 & \frac{\Delta^2}{2} \\
                                \Delta & 0 \\
                                0 & \frac{\Delta}{\tilde{v}}\end{bmatrix}.
    \end{aligned}
\end{equation}
Notice that the admissible action space $\mathcal{U}$ of a general vehicle is dependent upon various factors. While we assume the action space remains the same throughout the $T$ steps of propagation for safety evaluation, the action space $\mathcal{U}(v, \eta, t)$ is initialized as a function of the vehicle type $\eta$ and the current velocity $v$. Fig.~\ref{fig:a space veh} shows the maximum longitudinal acceleration $a_x^{\max}$ as an approximated polynomial function of $v$ and $\eta$ for $v \geq 0$. The data is collected and analyzed from real tests with a selected set of 6 representative vehicles. These vehicles consisted of 3 battery-electric vehicles and 3 combustion engine powered vehicles since the two powertrains demonstrate significantly different acceleration profiles. Different acceleration profiles represent a difference in control capabilities, which may lead to different MPrISM scores for the same traffic snapshot. The lateral acceleration bound $a_y^{\min}, a_y^{\max}$ and the maximum deceleration $a_x^{\min}$ are not shown in Fig.~\ref{fig:a space veh}, but are also determined with respect to $v$ and $\eta$. We also approximate the ellipse acceleration map with a dodecagon following the similar method from~\cite{junietz2018criticality} (partially illustrated within one quadrant in Fig.~\ref{fig:a space veh}). This can be formulated as 12 inequality constraints as follows for the action taken at $T=1$ in the optimization problem of~\eqref{eq: minmax opt}. Notice that the lateral action bound satisfies $|a_{y}^{\min}| = |a_{y}^{\max}|$.
\begin{equation}
    \label{eq: dodecagon}
    \begin{aligned}
    &\begin{bmatrix}
    \mathbf{L}_x^{\min} & \mathbf{L}_y^{\max} \\
    \mathbf{L}_x^{\min} & -\mathbf{L}_y^{\max} \\
    -\mathbf{L}_x^{\max} & \mathbf{L}_y^{\max} \\
    -\mathbf{L}_x^{\max} & -\mathbf{L}_y^{\max} \\
    \end{bmatrix} \mathbf{u} \leq \mathbf{b}_{xy},
    \mathbf{b}_{xy} = \begin{bmatrix} \sin{\frac{5\pi}{12}} \\ \ldots \\ \sin{\frac{5\pi}{12}} \end{bmatrix}, \\
    & \mathbf{L}_x^{\min} = \begin{bmatrix}  
    \frac{6}{5|a_{x}^{\min}|} \cos{\frac{7\pi}{12}} \\
    \frac{6}{5|a_{x}^{\min}|} \cos{\frac{9\pi}{12}} \\
    \frac{6}{5|a_{x}^{\min}|} \cos{\frac{11\pi}{12}} 
    \end{bmatrix},
    \mathbf{L}_y^{\max} = \begin{bmatrix}  
    \frac{1}{|a_{y}^{\max}|} \sin{\frac{7\pi}{12}} \\
    \frac{1}{|a_{y}^{\max}|} \sin{\frac{9\pi}{12}} \\
    \frac{1}{|a_{y}^{\max}|} \sin{\frac{11\pi}{12}} 
    \end{bmatrix}.
    \end{aligned}
\end{equation}

We emphasize that the admissible action space can be easily extended to include other constraints with respect to various factors of traffic rules, speed limit, ethical concerns, responsibility, etc. 

\subsubsection{Pedestrian kinematics}
We adopt the Dubins-car-like model with mild modifications in approximating the kinematics of the pedestrian as follows.
\begin{equation}
    \label{eq: ped}
    \dot{\mathbf{x}} = \begin{bmatrix} \dot{p} \\ \dot{q} \\ \dot{\phi} \end{bmatrix} = \begin{bmatrix} v\cos{(\phi)} \\ v\sin{(\phi)} \\ \gamma \end{bmatrix}.
\end{equation}
The pedestrian can reach a instantaneous velocity of $v \in [0, 3 m/s]$ and heading angle rate of $\gamma \in [-0.3 rad/s, 0.3 rad/s]$. Assume small course angle change, we can linearize the model and numerically approximate the solution as
\begin{equation}
    \label{eq: linear ped rk4}
    \begin{aligned}
        & \mathbf{x}' = \mathbf{A}_p \mathbf{x}  + \mathbf{B}_p \mathbf{u} , \mathbf{x} = [p, q, \phi]^T, \mathbf{u}=[v, \gamma]^T \\
        & \mathbf{A}_p = \begin{bmatrix} 1 & 0 & 0 \\
                                0 & 1 & 0 \\
                                0 & 0 & 1 \end{bmatrix},
          \mathbf{B}_p = \begin{bmatrix} \Delta & 0 \\
                                0 & \frac{\tilde{v} \Delta^2}{2} \\
                                0 & \Delta \end{bmatrix}
    \end{aligned}
\end{equation}

While we only study the motion of vehicles and pedestrians in this section, the method can easily generalize to a broad category of traffic objects through altering the admissible action space. 

\subsection{The Minimax Quadratic Optimization Formulation}
For a multi-agent system of various vehicles and pedestrians where a certain vehicle is selected as the SV, we can rewrite the target function $h_i(\mathbf{x},\mathbf{u},t+T\Delta)$ for the optimization problem of~\eqref{eq: minmax opt} as
\begin{equation}
    \label{eq: qp target}
        \norm{ \hat{\mathbf{A}_i} \mathbf{x}_i(t) + \hat{\mathbf{B}_i} \{\mathbf{u}_i(t)\} - \hat{\mathbf{A}}_0 \mathbf{x}_0(t) - \hat{\mathbf{B}}_0 \{\mathbf{u}_0(t)\}}_2,
\end{equation}
where $\hat{\mathbf{A}} = \left( \mathbf{A} \right)^T$ and $\hat{\mathbf{B}} = \begin{bmatrix} \mathbf{A}^{T-1}\mathbf{B} & \ldots & \mathbf{AB} & \mathbf{B} \end{bmatrix}$. $\mathbf{\hat{A}} \in \mathbb{R}^{N \times N}$ and $\hat{\mathbf{B}} \in \mathbb{R}^{N \times T \times M}$. This can be further propagated to the form of
\begin{equation}
    \label{eq: qp minimax}
    \begin{aligned}
    &\underset{\bar{\mathbf{u}}_i}{\min}\underset{\bar{\mathbf{u}}_0}{\max} \ 
    \bar{\mathbf{u}}_i^T\mathbf{P}\bar{\mathbf{u}}_i\!+\!\bar{\mathbf{u}}_0^T\mathbf{Q} \bar{\mathbf{u}}_0\!
    +\!\bar{\mathbf{u}}_i^T\mathbf{R}\bar{\mathbf{u}}_0\!+\!\mathbf{U}^T \bar{\mathbf{u}}_i\!
    +\!\mathbf{V}^T \bar{\mathbf{u}}_0\!+\!H \\
    & \text{subject to}\ \ \ \ \ \ \ \ \ \ \ \ \mathbf{L}_i\bar{\mathbf{u}}_i \leq \mathbf{b}_i,\mathbf{L}_0\bar{\mathbf{u}}_0 \leq \mathbf{b}_0,
    \end{aligned}
\end{equation}
where $\mathbf{P}$ and $\mathbf{Q}$ are positive definite. The inequality constraints are mainly determined by the action space of the participating agents. We can also extend the constraints considering the drivable space, traffic rules, responsibility, etc.

To this end, we have reduced the MPrISM algorithm to solving a series of the minimax quadratic optimization problems of~\eqref{eq: qp minimax}. Let the optimization target be $J(\cdot)$, \eqref{eq: qp minimax} can be solved with alternating gradient descent (AGD) as
\begin{equation}
    \label{eq: alternating gd}
    \bar{\mathbf{u}}_i \leftarrow \bar{\mathbf{u}}_i - \rho \triangledown_{\bar{\mathbf{u}}_i} J(\cdot), \text{and} \ \bar{\mathbf{u}}_0 \leftarrow \bar{\mathbf{u}}_0 + \mu \triangledown_{\bar{\mathbf{u}}_0} J(\cdot),
\end{equation}
with projection handling the constraints conditions. Generally, given the maximization target is convex,~\eqref{eq: qp minimax} is essentially non-convex optimization, AGD only guarantees the convergence to a local optimal solution. However, the quadratic form in~\eqref{eq: qp target} is of a particular structure that would provide efficient global optimal convergence with tractable solutions. 

Let the critical point be $\bar{\mathbf{u}}^* := [\bar{\mathbf{u}}_0^*,\bar{\mathbf{u}}_i^*] = \text{argmin}_{\bar{u}} J(\cdot)$, 
and the convex set induced by the linear inequalities be $\Lambda := \{ \bar{\mathbf{u}} | \mathbf{L}_i\bar{\mathbf{u}}_i \leq \mathbf{b}_i,\mathbf{L}_0\bar{\mathbf{u}}_0 \leq \mathbf{b}_0 \}$. Consider the following two cases.

Case 1: If $\bar{\mathbf{u}}^* \notin \Lambda$,~\eqref{eq: minmax opt} becomes convex optimization and AGD guarantees the convergence to the global optimum. In the specific practice of an ADS safety evaluation, this implies that the action from the traffic object to initiate the worst-case scenario and the optimal response from the SV to escape from such a worst-case both lie on the boundary of each agent's action profile. Most of the cases one would encounter during the ADS safety evaluation satisfy such a condition hence can be solved efficiently with AGD. 

Case 2: On the other hand, if $\bar{\mathbf{u}}^* \in \Lambda$, the optimization becomes non-convex. While the optimal action trajectory for the SV still lies on the boundary of the action profile, it requires considering the full control authority of the traffic object to determine the optimal attacking trajectory. In practice, the number of such cases is limited, given the condition requires a relatively close distance between the two participants. While slower than AGD in the convex setting, a tractable solution can still be obtained through the branch-and-bound technique that guarantees the convergence to the global optimum. 

One can also adopt various accelerated gradient descent methods~\cite{nesterov2013introductory} to facilitate optimization.
Notice that Algorithm~\ref{alg:ttc} also naturally supports parallel computation. 
In practice, the algorithm is efficient for various applications of ADS safety evaluation. Details are described in Section~\ref{sec:examples}.

\section{Examples}
\label{sec:examples}
The implementation of the proposed MPrISM metric for ADS safety evaluation is empirically illustrated with three types of examples, the snapshot, the scenario, and the on-road test. The snapshot is provided as a fixed time instance. A scenario describes a testing procedure for automated vehicles. The SV is deployed in a contained environment with a set of traffic objects following the pre-determined or conditional motion trajectories. The on-road test further extends the scenario to bigger maps with more complex flows of traffic. 

\begin{figure}[t]
    \centering
    \vspace{0.2cm}
    \includegraphics[width=0.45\textwidth]{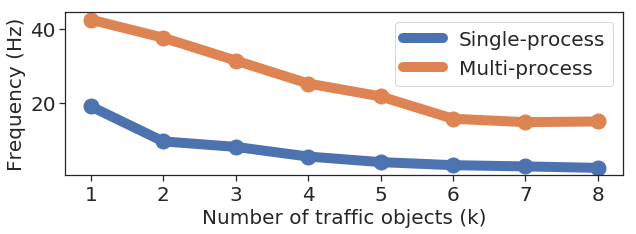}
    \caption{Running frequency of MPrISM with respect to the number of traffic objects empirically evaluated using single-process and multi-process (8 workers)}
    \label{fig:freq}
\end{figure}

Throughout the examples shown in this section, we use the same set of MPrISM parameters of $C=2, \Delta=0.1$, $T=10$. That is, we consider the collision as a circled region of 2$m$ in radius with respect to a certain selected center of the SV. Furthermore, we have $\tau \in \{0, 0.1, \ldots, 1 \}$, propagating the time to collision within the one-second look-ahead time period ($\tau=1$ if it is collision-free). We implement the algorithm in Python. With the Intel Xeon 3.4GHz CPU core, Fig.~\ref{fig:freq} illustrates the empirical performance of MPrISM for single-process and multi-process (8 workers) given repeatedly sampled snapshots with various number of traffic objects. In practice, a traffic scene leading to a certain fatal accident typically involves no more than 5 participants~\cite{forkenbrock2015nhtsa}. This leads to a processing rate of over 20 snapshots per second with 5 traffic objects in each snapshot under multi-process.

\subsection{Snapshots}
\begin{figure}
    \centering
    \vspace{0.2cm}
    \includegraphics[width=0.45\textwidth]{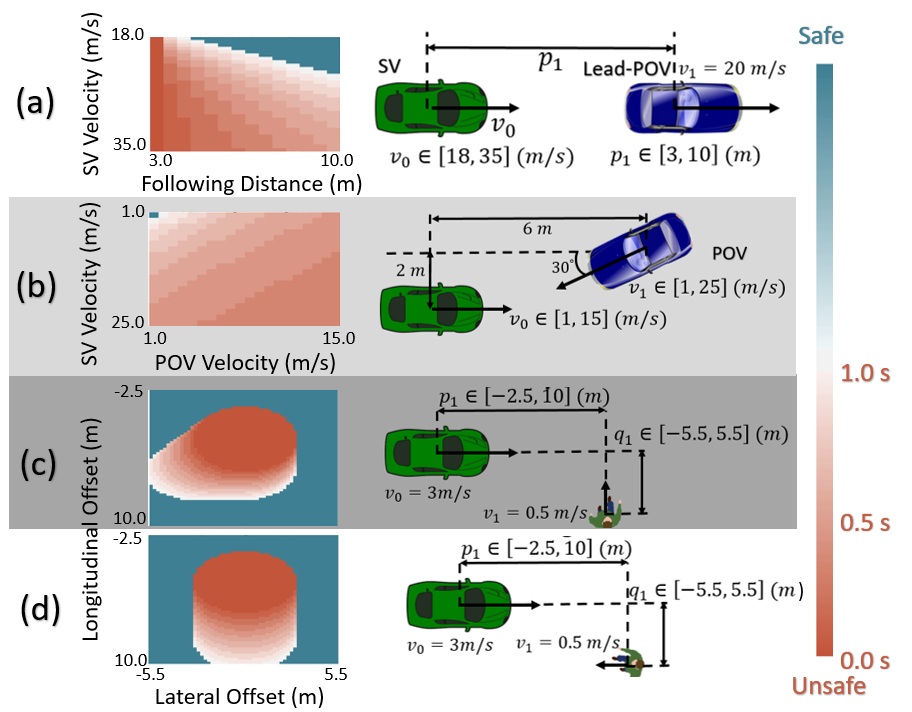}
    \caption{Various snapshots safety analysis: (a) The "Lead-POV Following" Snapshots, (b) The "Oncoming Left-turning POV" Snapshots, (c) The "Perpendicularly Passing Pedestrian" Snapshots, (d) The "Oncoming Pedestrian" Snapshots}
    \label{fig:snapshots_safety}
\end{figure}

We assume identical vehicles for all snapshots studied in this section. Fig.~\ref{fig:snapshots_safety} illustrates the MPrISM analysis results of vehicle-to-vehicle and vehicle-to-pedestrian interactions with respect to a selected pair of variables.

For the Lead-POV Following snapshots (Fig.~\ref{fig:snapshots_safety}(a)), SV is safe when it is traveling at a lower velocity than the lead-POV. As SV increases its speed along with the lead-POV decreasing the speed, the snapshot becomes less safe (smaller MPrTTC). In the Oncoming Left-turning POV cases (Fig.~\ref{fig:snapshots_safety}(b)), SV is safe when it has a higher velocity than the slower moving rear POV. The situation becomes less safe with decreasing SV velocity and increasing rear-POV velocity. Given the configuration of the Perpendicularly Passing Pedestrian snapshots (Fig.~\ref{fig:snapshots_safety}(c)), the pedestrian is more likely to initiate a potential collision from the right side of the vehicle. On the other side, the pedestrian holds equal capability to cause a potential collision on both sides of the SV in the analysis of the Oncoming Pedestrian snapshots.

\begin{figure}
    \centering
    \vspace{0.2cm}
    \includegraphics[width=0.45\textwidth]{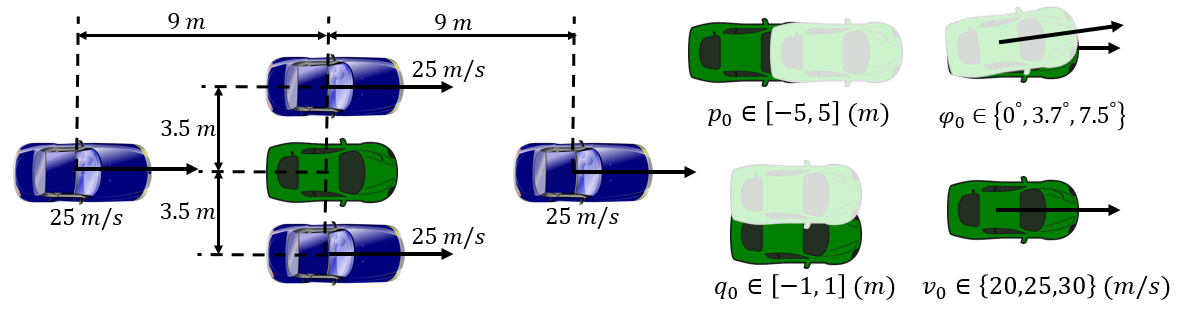}
    \caption{The boxed-in snapshots with four controlled variables.}
    \label{fig:boxedin snap}
\end{figure}

\begin{figure}
    \centering
    \vspace{0.2cm}
    \includegraphics[width=0.49\textwidth]{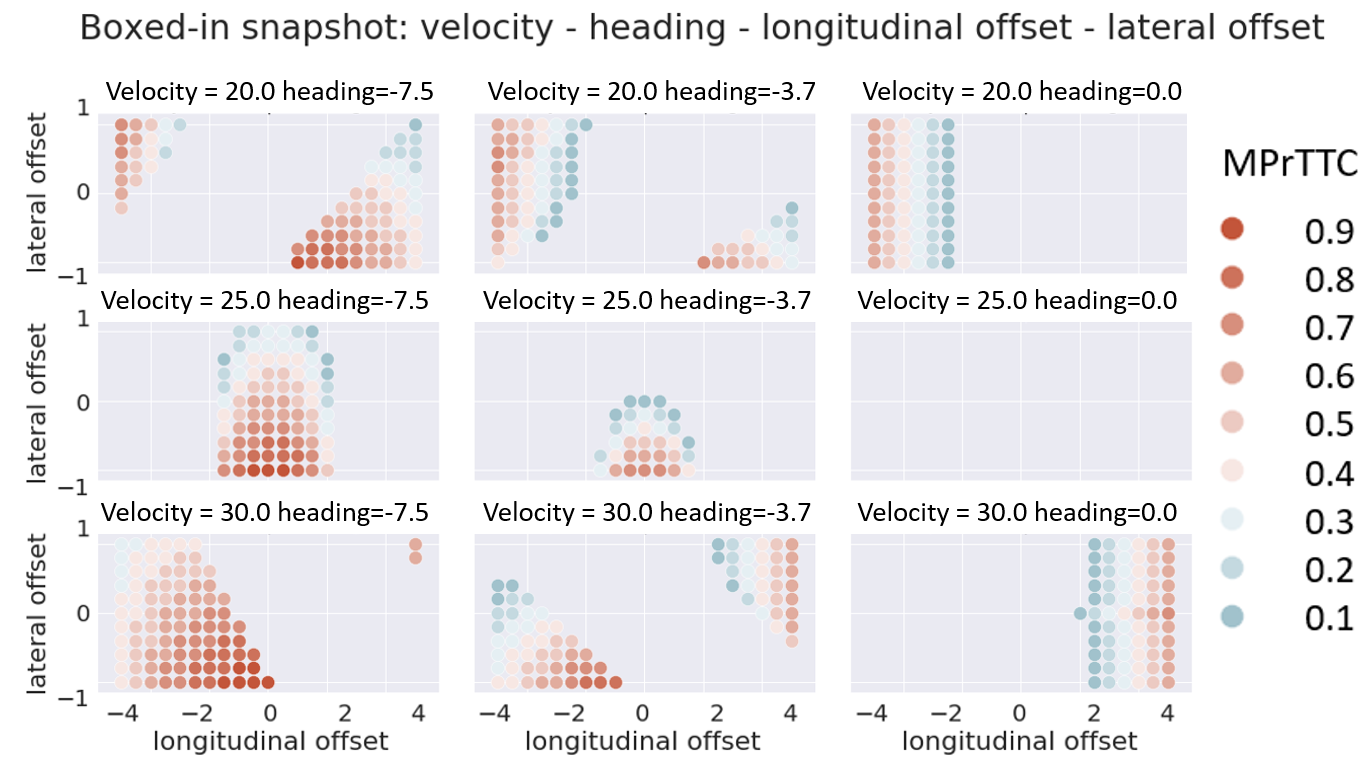}
    \caption{The boxed-in safety analysis with unsafe cases only.}
    \label{fig:boxedin safety}
\end{figure}

We also study a group of snapshots in the boxed-in scenario where the SV is surrounded by 4 identical traffic vehicles (Fig.~\ref{fig:boxedin snap}). We study the MPrISM results with respect to four SV controlled variables, including the longitudinal offset, the lateral offset, the velocity, and the heading angle. We perform MPrISM analysis on a total of 2079 snapshots among which we have 571 snapshots considered unsafe (Fig.~\ref{fig:boxedin safety}). Following the partially non-cooperative assumption of~\ref{asp:partial non-coop} and Algorithm~\ref{alg:ttc}, one can derive the specific traffic object that dominates the crash through $\text{argmin}_{i=1, \ldots, k} \{ \tau_i \}$. Within the 571 unsafe snapshots, the front-POV, the rear-POV and the left-POV dominate 101, 140 and 330 potential collisions, respectively. Notice that we only set the choices of heading angle to the left side of the SV.

When the SV matches velocity with the group of POVs at $25 m/s$ with the same heading angle, the situation is safe regardless of the choices of positioning offset, leading to an empty subplot as shown by the far-right column, middle row in Fig.~\ref{fig:boxedin safety}. SV tends to experience more unsafe snapshots at higher velocity ($30 m/s$), as well as a higher deviation from the zero heading angle ($-7.5^{\circ}$).

Overall, the MPrISM analysis of various snapshots align with the general intuitive notion of safe driving behaviors, and the numeric results provide objective details that would enhance our understanding regarding the safety evaluation of ADS. The capability of MPrISM to serve as a safety scoring system in practice is shown in the following examples of traffic scenarios.

\subsection{Traffic Scenarios}
\begin{figure}[b]
    \centering
    \includegraphics[width=0.45\textwidth]{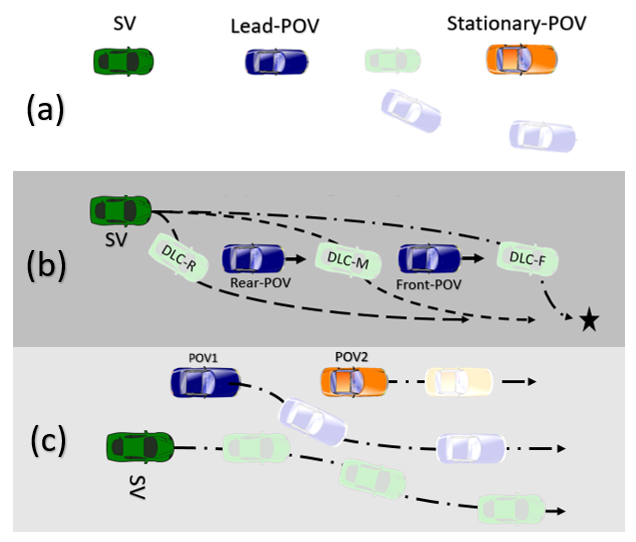}
    \caption{Various traffic scenarios configurations: (a) The "suddenly revealed stopped vehicle" (SRSV), (b) The "highway exit" and (c) The "emergent single lane change".}
    \label{fig:scenes}
\end{figure}

Three traffic scenarios are studied in this section (Fig.~\ref{fig:scenes}(a)). The "suddenly revealed stopped vehicle" (SRSV) is a standard testing scenario with data collected from real-world tests. The "highway exit" and the "emergent single lane change" are deployed in the VIRES VTD simulator. The admissible action space is customized with respect to the type of vehicle involved.

\subsubsection{The suddenly revealed stopped vehicle}
\begin{figure}
    \centering
    \vspace{0.2cm}
    \includegraphics[trim=0 13.5cm 0 0, clip,width=0.45\textwidth]{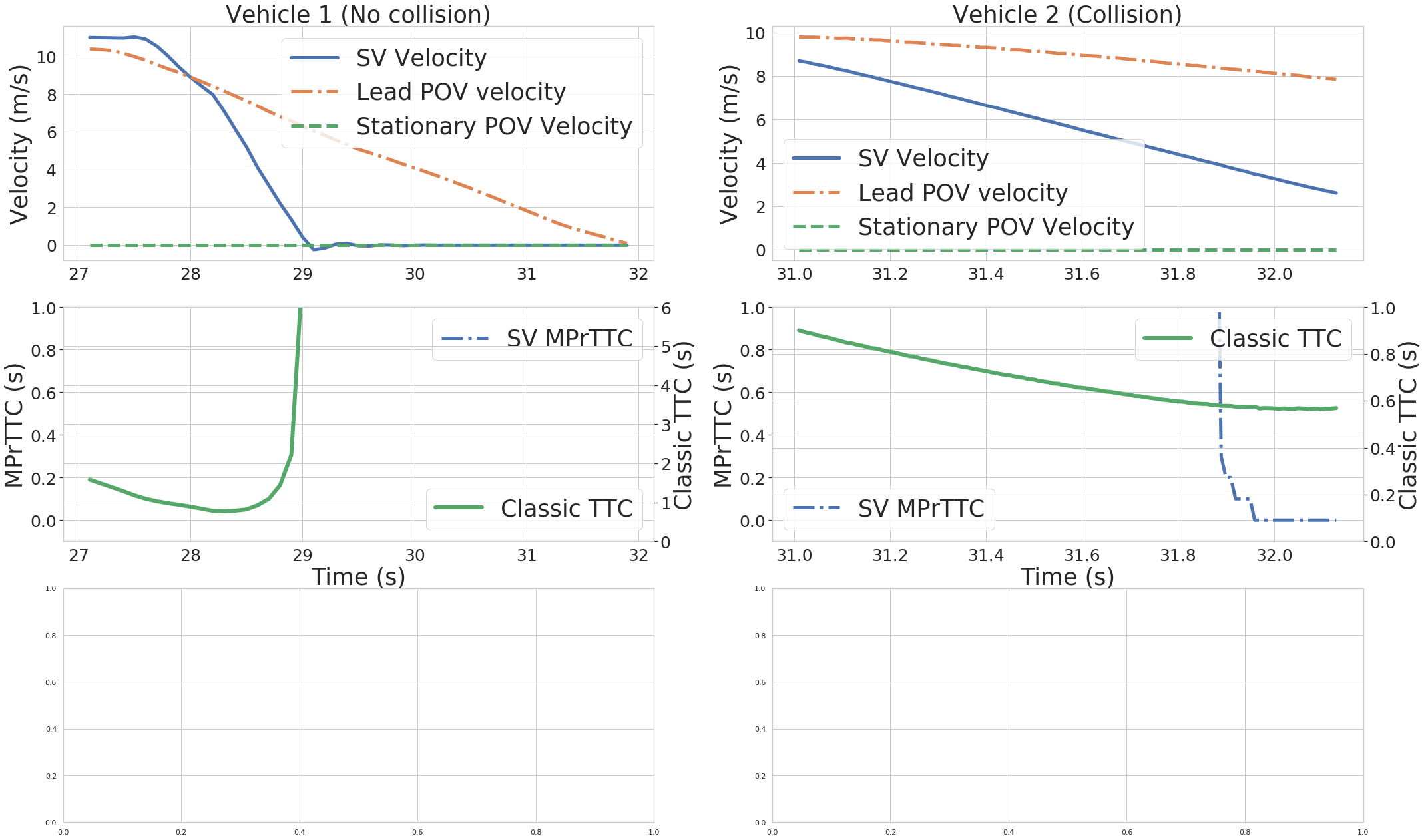}
    \caption{MPrISM analysis comparing two vehicles in the SRSV scenario where Vehicle 2 ends up with a collision.}
    \label{fig:srsv safety}
\end{figure}

The objective of the SRSV test is to evaluate the traffic jam assist (TJA) system’s performance in detecting and properly responding to a stationary POV that is suddenly revealed after another lead POV steers to the side. All data analyzed in this scenario was collected from real vehicle tests on two commercially available vehicles equipped with the TJA system. Vehicle 2 (Fig.~\ref{fig:srsv safety}) failed the test with a collision. For both vehicles, MPrISM analysis shows MPrTTC values greater than 1 throughout the first stage when SV is following the lead-POV. The comparison between two vehicles in the brake-to-stop stage is shown in Fig.~\ref{fig:srsv safety}. Vehicle 1 manages to brake to a stop without a collision. This aligns with the MPrISM analysis where no MPrTTC results are plotted, indicating that no collision would occur within the 1-second imminent future from any time step throughout the test. For Vehicle 2, the MPrTTC turns to $0.2$ at $0.23s$ before the actual collision, indicating an MPrTTC of $\tau=0.2s$ for the SV when presented with the worst-case scenario.
Currently, NCAP~\cite{van2017euro} like testing procedures and scenarios are designed for validating low-level algorithms. Test data collected in these tests can be partially analyzed by observing collision results and applying the classic TTC metric. The following scenarios are provided to show the performance of MPrISM in more complicated tests regarding evaluating high-level planning and control algorithms.

\subsubsection{The highway exit}
This scenario takes place on a three-lane highway. The SV begins in the left-most lane and is trying to navigate to the highway exit on the right lane through two POVS in the middle lane. Generally, the path planning algorithm would select one of the three possible trajectories of performing the dual-lane change maneuver; pass in the front of (DLC-F), through the middle (DLC-M), or on the rear (DLC-R) of the two POVs. With the appropriate setup, all of the options could get through the test with acceptable performance in terms of utility (reaching the exit on time without breaking traffic rules or causing any collisions). Here we analyze them using MPrISM.
\begin{figure}
    \centering
    \vspace{0.2cm}
    \includegraphics[width=0.45\textwidth]{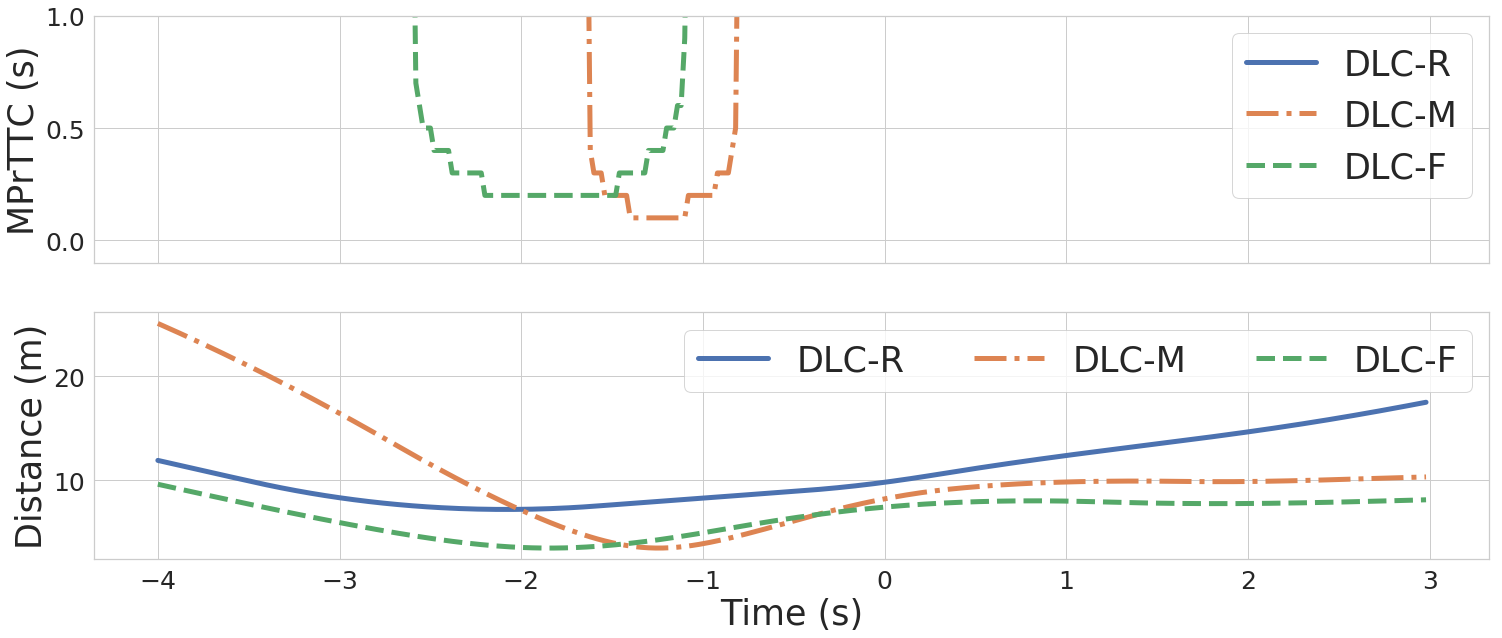}
    \caption{Safety analysis of the highway exit scenario, the distance shown in the second plot is measured as the relative lateral distance between the SV and the laterally nearest POV.}
    \label{fig:hwe safety}
\end{figure}
The scenario starts with the two POVs maintaining the same lane following behavior throughout the test. All vehicle states are recorded at 50\text{Hz}. For each test, we select the time when the SV reaches the center of the middle lane as $t=0$ and take all vehicle states within the time range of $[-4s, 3s]$. From Fig.~\ref{fig:hwe safety},it is shown that the DLC-R had an MPrTTC value greater than 1 for the entire scenario. Although both DLC-F and DLC-M achieved about the same minimum relative distance at the same maximum velocity, MPrISM is able to catch the geometric difference of the specific traffic snapshot, which shows that the DLC-M scenario results in smaller MPrTCC values.  These values, resulting from the MPrISM method, hold potential for safety evaluations. In this example, the MPrTCC values suggest DLC-R is safer than either DLC-F or DLC-M, with DLC-F slightly safer than DLC-M.  
.

\subsubsection{Emergent single lane change}
\begin{figure}
    \centering
    \vspace{0.2cm}
    \includegraphics[width=0.45\textwidth]{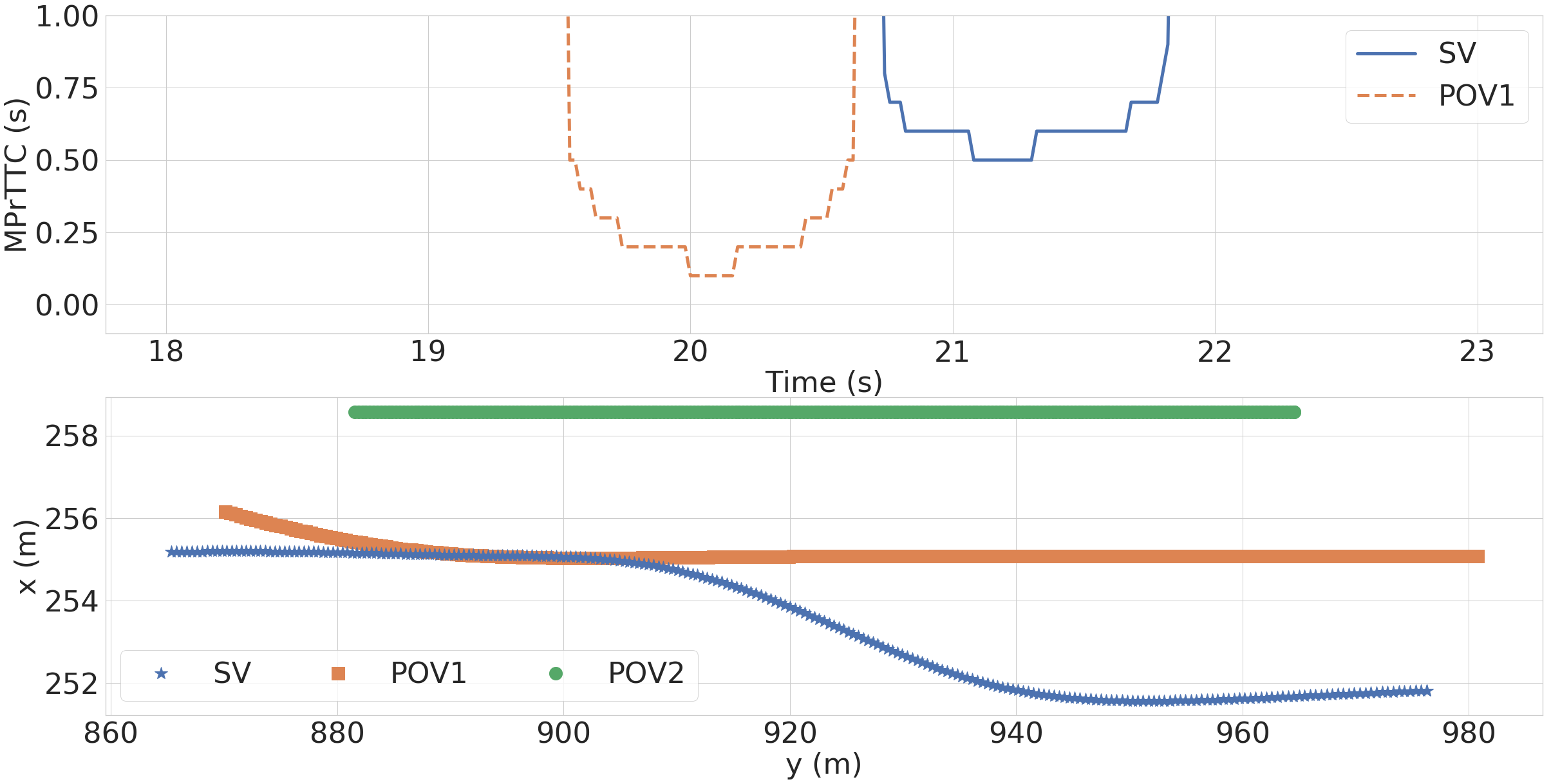}
    \caption{Safety analysis of SV and POV1 in the emergent single lane change scenario.}
    \label{fig:pegasus safety}
\end{figure}

The scenario of the emergent single lane change is shown in Fig.~\ref{fig:scenes}(c). A similar case was also studied in the previous work of~\cite{pegasus2018} from the criticality perspective. The MPrTTC and vehicle states are shown in Fig.~\ref{fig:pegasus safety}. In addition to studying safety from the SV perspective, we also consider the POV1 as the test subject with MPrISM analysis. It is interesting to observe that POV 1 has lower MPrTCC values than the SV vehicle, suggesting it experienced a more dangerous situation that the SV based on the MPrISM results. This also aligns with the general intuition that the POV1 is the more aggressive agent in this scenario. 

\subsection{On-road Driving Simulation}
Another venue for demonstrating MPrISM results is simulated on-road driving. For this, we simulate naturalistic on-road driving in SUMO~\cite{krajzewicz2002sumo} with two automated driving algorithms deployed on the same set of maps through the same traffic scenarios operating for the same amount of time. Table~\ref{tbl: on-road info} lists the details of the simulation setup. Driver 1 is designed to be more aggressive than the Driver 2. Throughout the 12000-second simulation, neither of the drivers was reported with collisions or disobeying traffic rules.


\begin{table}
\centering
\vspace{0.2cm}
\caption{Parameters of map, traffic, and automated driving algorithms for SUMO simulation of the on-road driving.}
\begin{tabular}{c | c || c | c | c }
\toprule
\multicolumn{2}{c}{SUMO setup} & \multicolumn{3}{c}{Driver setup} \\ \hline
Edges & 57614  & Model            & Driver 1     & Driver 2 \\
Lanes & 60930 & minGap           & 0.5          & 3   \\
Junctions & 86 & sigma            & 1            & 0.5 \\
Distance($km$) & 738.564  & tau              & 0.1          & 2   \\
3 Lane Edges & 158 & emergencyDecel   & 4            & 6  \\
2 Lane Edges & 2794 & lcStrategic      & $10^5$       & 1  \\
Vehicle flows & 18 & lcSpeedGain      & $10^6$       & 1  \\
Vehicles & 602 & lcKeepRight      & 100          & 1   \\
Pedestrian flows & 7 & lcOvertakeRight  & 100          & 1  \\
Step size & 0.02$s$ & lcOpposite       & 1000         & 1   \\
\ & \ & lcAssertive      & 1000         & 1         \\
\bottomrule
\end{tabular}
\label{tbl: on-road info}
\end{table}

\begin{table}[h]
\centering
\caption{Safety comparison of two intelligent drivers in the on-road test.}
\begin{tabular}{c | c | c }
\toprule
Model                                         & Driver 1                       & Driver 2   \\\hline
Total running time ($s$)                      &   \multicolumn{2}{c}{12000} \\
Total running distance ($km$)                  & 151.38                     & 161.6 \\ 
Average $a_x$ ($m/s^2$)              & -0.001                        & 0.004 \\ 
Average number of near traffic vehicles            & 1.641                         & 1.53  \\\hline
Total time of being unsafe ($s$)         & \textbf{111.64}                         & 96.4       \\
Mean $a_x$ when unsafe ($m/s^2$)     & \textbf{-0.258}                         & -0.064 \\
\bottomrule
\end{tabular}
\label{tbl: on-road safety}
\end{table}

From Table~\ref{tbl: on-road safety}, given that Driver 1 stays being unsafe for longer period of time then Driver 2. This aligns with the setup shown in Table~\ref{tbl: on-road info} where Driver 1 is designed to be more aggressive. Also, one interesting observation is that even though the overall acceleration of both drivers remain close, Driver 1 tends to perform more extreme maneuvers (deceleration with higher magnitude) when it is operating in the unsafe status judged by MPrISM.

\section{Conclusion}
\label{sec:conclusion}
A testing metric for planning and control performance evaluation of ADS is introduced. The metric determines the time to a collision of the subject vehicle, given the worst possible scenario with respect to various choices of actions. It objectively scores the performance of an automated driving system. The implementation of the MPrISM method is shown through numerical examples, simulated testing scenarios, and real-world vehicle tests. 

While the proposed metric is derived in a modeled manner with deterministic dynamic transitions and explicit actuation constraints, it is worth mentioning that the metric is still compatible with probability techniques for risk assessment. For example, one can impose a particular probability distribution over various action profiles and establish a weighted summary of the MPrISM results for each profile. However, this is out of the scope of this paper and is of interest for future work.  
Other future efforts could consider extending the target function of~\eqref{eq: qp target} with more factors. Some of such changes would potentially affect the nature of the optimization target, but could also make a more representative term for safety. 

\bibliographystyle{unsrt}  
\bibliography{references}

\end{document}